\documentclass{article}
\usepackage{ijcai24}
\usepackage{xspace} 
\usepackage{amsmath} 
\usepackage{amsfonts}
\usepackage{stfloats}
\usepackage[numbers]{natbib}
\makeatletter
\DeclareRobustCommand\onedot{\futurelet\@let@token\@onedot}
\def\@onedot{\ifx\@let@token.\else.\null\fi\xspace}
\def\eg{\emph{e.g}\onedot} 
\def\ie{\emph{i.e}\onedot}

\def\etal{\emph{et al}\onedot}
\makeatother
\usepackage{times}
\usepackage{soul}
\usepackage{url}
\usepackage[hidelinks]{hyperref}
\usepackage[utf8]{inputenc}
\usepackage[small]{caption}
\usepackage{graphicx}
\usepackage{amsmath}
\usepackage{amsthm}
\usepackage{booktabs}
\usepackage{algorithm}
\usepackage[switch]{lineno}
\usepackage{algpseudocode}
\usepackage{pifont}

\newtheorem{example}{Example}
\newtheorem{theorem}{Theorem}

\title{Brain Inspired Adaptive Memory Dual-Net for Few-Shot Image Classification}

\author{
Kexin Di$^1$
\and
Xiuxing Li$^2$\and
Yuyang Han$^1$\and
Ziyu Li$^2$\and
Qing Li$^2$\And
Xia Wu$^2$\\
\affiliations
$^1$Beijing Normal University\\
$^2$Beijing Institute of Technology\\
\emails
202321081053@mail.bnu.edu.cn
}

\begin{document}

\maketitle

\begin{abstract}
    Few-shot image classification has become a popular research topic for its wide application in real-world scenarios, however the problem of supervision collapse induced by single image-level annotation remains a major challenge. Existing methods aim to tackle this problem by locating and aligning relevant local features. However, the high intra-class variability in real-world images poses significant challenges in locating semantically relevant local regions under few-shot settings. Drawing inspiration from the human's complementary learning system, which excels at rapidly capturing and integrating semantic features from limited examples, we propose the generalization-optimized Systems Consolidation Adaptive Memory Dual-Network (\textbf{SCAM-Net}). This approach simulates the systems consolidation of complementary learning system with an adaptive memory module, which successfully addresses the difficulty of identifying meaningful features in few-shot scenarios. Specifically, we construct a Hippocampus-Neocortex dual-network that consolidates structured representation of each category, the structured representation is then stored and adaptively regulated following the `generalization optimization' principle in a long-term memory inside Neocortex. Extensive experiments on benchmark datasets show that the proposed model has achieved state-of-the-art performance. 
\end{abstract}

\section{Introduction}

\begin{figure}
  \centering
  \includegraphics[width=\linewidth]{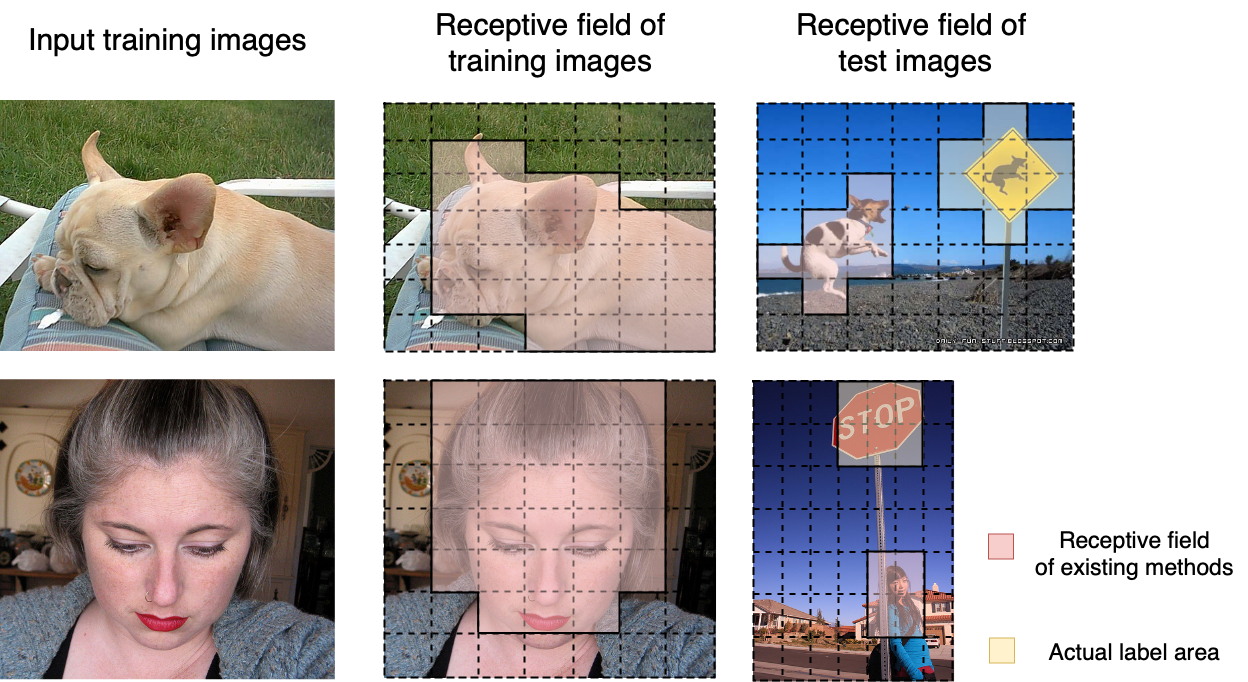}
  \caption{The label annotated `sign' but recognized as `dog' or `woman', meaning patterns that irrelevant to the test class but seen in training set are overemphasized. SCAM-Net aims to only consolidate the useful patches and ignore irrelevant elements as much as possible. \textit{Image source: miniImagenet} \cite{vinyals2016matching} }
  \label{fig:intro}
\end{figure}

Images that depict real-world scenes often contain multiple complex elements. However, many computer vision datasets, such as ImageNet \cite{russakovsky2015imagenet}, offer single-label annotations that only describe part of the image's content. While adequate for conventional classification tasks with ample data, the single-labeled datasets present challenges for few-shot learning (FSL). FSL aims to classify new categories that different from those in the training set with only a small number of examples (\eg, one or five). In this scenario, entities that are unrelated to the training classes but exist in the test set may be overlooked, while patterns that were important during training but are irrelevant to the test classes might be overemphasized, result in a phenomenon referred to as `supervision collapse' \cite{doersch2020crosstransformers}, as illustrated in Figure \ref{fig:intro}.

Dense feature-based methods have emerged as a promising approach to address interference and the supervision collapse issue in FSL. DeepEMD \cite{zhang2020deepemd} adopts Earth Mover’s Distance as a metric to compute dense image representations to determine image relevance. FewTURE \cite{hiller2022rethinking} employs a reweighing strategy to identify the most informative region. CPEA \cite{hao2023class} integrates patch embeddings with class-aware embeddings to make them class-relevant thus avoiding the need to align semantically relevant regions. However, the significant intra-class variability and background clutter can lead to distinct distribution between source and target data, commonly referred to as `data drift'. This variability presents challenges when locating semantically relevant regions, thereby affects the generalization ability of FSL models.

Given the limitations in previous studies, we aim to tackle the aforementioned issues from a biological perspective. Unlike machine intelligence, human brain can rapidly capture and integrate semantic features from limited examples in data drifting conditions. Related cognitive functions rely on \text{systems consolidation} mechanisms that construct neocortical memory traces from hippocampal precursors \cite{mcclelland1995there}. The Hippocampus and the Neocortex is a set of complementary learning system, Hippocampus exhibits short-term adaptation and rapid learning of episodic information which is then gradually consolidated to the Neocortex for structured information \cite{schapiro2017complementary,atallah2004Hippocampus}. Besides this basic consolidation process, a recent research has shown that unregulated neocortical memory transfer can lead to over-fitting and harm generalization in new environments \cite{sun2023organizing}, therefore memory is consolidated only when it aids generalization, which is called \textbf{generalization-optimized systems consolidation}. The principle provides us insight into how adaptive behavior benefits from complementary learning systems specialized for memorization and generalization. 

In this paper, we propose a \textbf{generalization-optimized Systems Consolidation Adaptive Memory dual-Network, SCAM-Net}. This design enhances the semantic feature representations and introduces an adaptively adjusted memory mechanism to further improve the representation reliability. Specifically, we construct a Hippocampus-Neocortex dual network that consolidates structured information through an adaptively regulated long-term memory. The Neocortex model integrates both spatial and semantic feature together enhances the representation ability of Hippocampus model, the Hippocampus model then constraints the Neocortex model by a slow update of exponential moving average (EMA) of the Hippocampus network weights.

The contributions of this work can be summarized as follows,
\begin{itemize}
 \item To address the issues of locating semantically relevant feature in traditional few-shot methods, we propose an adaptive memory dual-network inspired by the human brain's generalization-optimized systems consolidation mechanism. To the best of our knowledge, we are the first to incorporate the principle of optimize generalization into a systems consolidation dual network.
 \item We introduce SCAM-Net, a technique that: 1) enhances semantic feature representation through a hippocampus-neocortex dual model that consolidates structured information; and 2) adaptively refines this structured information using prior knowledge aligned with the optimized generalization principle. 

 \item We have conducted extensive experiments and achieved state-of-the-art performance on four benchmark datasets. The ablation study and visualization of adaptive memory regulation proved our superior advantage on miniImagenet dataset.
\end{itemize}

\section{Related Work}

\textbf{Few-shot Image Classification.}
Few-shot image classification has attracted considerable attention for its practical use in real-world scenarios. Approaches in this field can generally be categorized into two main streams: optimization-based methods and metric-based methods. Optimization-based methods focus on training a meta-learner that efficiently tunes a base-learner using a limited number of labeled images. Such examples include MAML \cite{finn2017model} and Reptile \cite{nichol2018first}, which provide a set of initial model parameters that can be quickly adapted to new tasks using a few gradient updates. Metric-based methods aim to develop a feature space (prototype) that is universally applicable across various tasks, using suitable distance metrics for similarity measurements to perform prototype-query matching. MatchingNet \cite{vinyals2016matching} utilizes neural networks to create a feature space where cosine distance measures similarity, while ProtoNet \cite{snell2017prototypical} employs Euclidean distance for the same purpose. Recently, dense feature-based methods have shown promising results in tackling the challenging few-shot image classification problem. One approach treats local features as the primary image representations \cite{sung2018learning,liu2022learning,xu2021learning,hao2023class,wu2024ammd}, while another aligns semantically relevant local features \cite{hou2019cross,zhang2020deepemd,doersch2020crosstransformers,hiller2022rethinking,leng2024meta}. 

The phenomenon of data drift in few-shot tasks presents challenges in feature representation and locating semantically relevant regions for existing methods \cite{luo2023closer}. Our approach takes into account both the spatial and structural feature of images, the structural feature is adaptively regulated based on prior knowledge. This allows for a more reliable semantically relevant feature representation for support samples, addressing the issues raised by significant intra-class variability and background clutter.

\begin{figure*}[h]
  \centering
  \includegraphics[width=\linewidth]{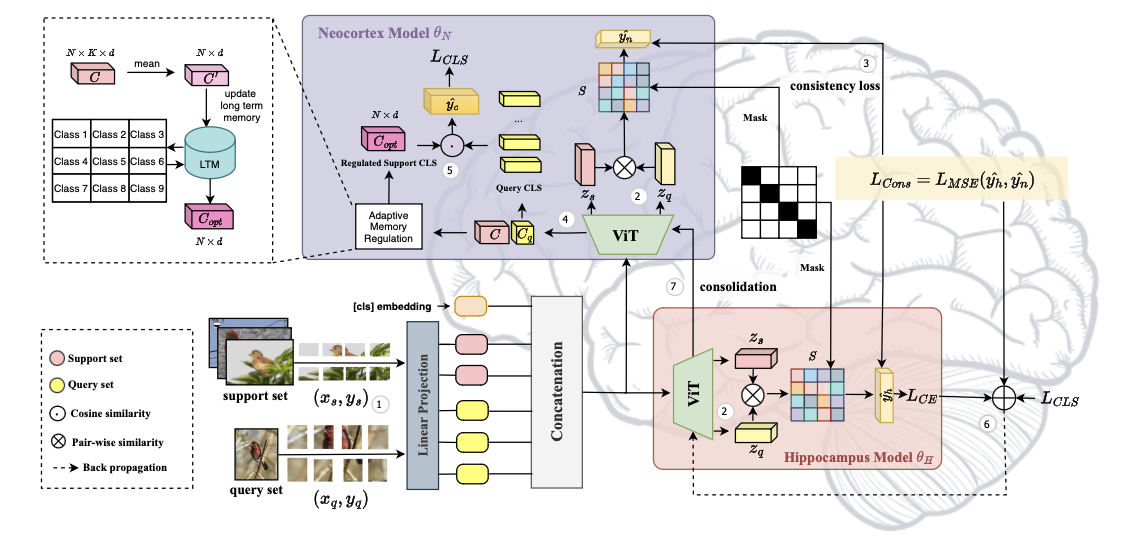}
  \caption{Overall architecture of the proposed SCAM-Net. \ding{172} After obtaining the patches of support set and query set, a [cls] embedding was concatenated with projected patches as input to both the Neocortex model and Hippocampus model. \ding{173} Inside the Neocortex model and Hippocampus model, the calculation of the similarity matrix between the patch embedding of support and query set embeddings obtained by ViTs is performed in parallel, block-diagonal masking is employed to prevent classifying the image itself. \ding{174} Based on which, we can get the prediction logits of Neocortex model and Hippocampus model and the consistency loss $L_{Cons}$ is calculated. \ding{175} The Neocortex model maintains a long-term memory that is auto-regulated due to each task. \ding{176} The regulated support CLS is obtained from Systems consolidation, and construct a correlation between the query CLS, sequently we obtain the loss $L_{cls}$. \ding{177}Afterwards, $L_{Cons}$ and $L_{cls}$ plus $L_{CE}$ together update the Hippocampus model. \ding{178} the Neocortex model perform consolidation to update.}
  \label{fig:framework}
\end{figure*}

\textbf{Brain-inspired FSL.}
With the recent advancements in brain science and computational neuroscience technology, several studies have sought to integrate brain mechanisms with few-shot learning to enhance the performance, achieving great success in application fields such as few-shot object detection and continual learning. Duan \etal \cite{duan2024few} designed a variational structured memory module (VSM) that promote few-shot generation task via recalling brain-inspired episodic-semantic memory. Sorscher \etal \cite{sorscher2022neural} proposes a neural mechanism inspired by the brain's ability to represent concepts as tightly defined manifolds. Complementary learning system was the inspiration for dual memory learning systems in earlier works \cite{french1999catastrophic}, and in recent years has shown it's potential in continual learning \cite{aranilearning,rostami2019complementary,han2024novel}, the studies that inspired by the biological model of fast/slow synapse effectively reduced catastrophic forgetting through memory mechanisms. However, these studies reveals an overlooked tension that unregulated Neocortex memory transfer can harm generalization \cite{spens2024generative,sun2023organizing}. In situations where limited reference samples provided, it becomes crucial to adaptively integrate and regulate memories base on the specific tasks. 

By incorporating the systems consolidation of complementary learning system, our method proposed a novel interaction between Hippocampus and Neocortex model, enabling the model to learn from structural information in the long-term memory. On the basis of systems consolidation, we are the first research to consider optimize generalization principle, that memories are regulated to further enhance feature representation and thereby alleviate the issue brought by data drift.

\section{Method}

\subsection{Problem Definition}
Inductive N-way K-shot few-shot image classification aims to adapt knowledge learned from training classes $C_{train}$ to unseen test classes $C_{test}$ with a limited number of labeled samples, $C_{train} \cap C_{test} = \emptyset$. We follow the meta-learning protocols established in prior studies \cite{vinyals2016matching} to formulate the episodic training and testing of few-shot classification problem. Each episode consists of a support set and a query set. The support set, $\mathcal{S} = \{(x_{s}^{i}, y_{s}^{i}) | i = 1, \dots, NK\}$, meaning $K$ images sample from $N$ different classes, and the query set, $\mathcal{Q} = \{(x_{q}^{i}, y_{q}^{i}) | i = 1, \dots, NQ\}$, where $Q$ denotes the number of test image in each class. The query set is used to evaluate the model's performance on few-shot classification task based on the support set. The objective is to train a model that can effectively generalize to episodes randomly sampled from the unseen test classes. 


\subsection{Overview}
Figure \ref{fig:framework}  shows the overall architecture of our SCAM-Net. In this work, we designed two major functional structure to alleviate supervision collapse: the Hippocampus-Neocortex dual-network with a novel interaction (detailed in Section 3.3) and an adaptive-regulated long-term memory inside Neocortex (detailed in Section 3.4). 
First of all, the support set and query set are encoded and form a similarity matrix in parallel in both of the Neocortex and Hippocampus model, result in prediction logits $\hat{y_n}$ and $\hat{y_h}$. The two models interact to enforce a consistency loss on the Hippocampus model which prevents rapid changes in the parameter space and as a guidance of semantic information from Neocortex. In addition, we take images' global representation into account to make compensation for inaccurately located dense feature. which adaptively regulated based on prior knowledge in the long-term memory to alleviate the negative effect of significant intra-class variability. 
The loss for Hippocampus model update consists of three components: the consistency loss, the loss of global representation of each category and cross entropy loss of Hippocampus model itself. The Hippocampus model constraints the Neocortex model through exponential moving average. The interplay between the functionality of the Hippocampus and Neocortex is crucial for concurrently learning efficient semantic representations. The overall pipeline algorithm is described in Algorithm 1.

\begin{algorithm}[!ht]
\caption{Generalization-optimized Systems Consolidation Algorithm}
\begin{algorithmic}[1] \label{algo}
\algnotext{Require}
\Require Data stream $\mathcal{D}$, Learning rate $\eta$, CLS weight $\lambda$, Decay parameter $\alpha$

\State \textbf{Initialize:} $\theta_H = \theta_N;  M \leftarrow \emptyset$ 
\While{Training}
    \State $(x, y) \sim \mathcal{D}$
    \State $\hat{y_h} = f(x; \theta_H)$
    \State $\hat{y_n} = f(x; \theta_N)$
    \State $L_{Cons} = L_{MSE}(\hat{y_h}, \hat{y_n})$
    \If{$L \notin M$}
        \State $M[L] = C'$
    \Else
        \State $M[L] = \frac{1}{2} (M[L] + C')$
    \EndIf
    \State $C_{opt} \leftarrow M[L]$
    \State $\hat{y_c} = \text{softmax}(\text{sim}(C_{opt}, C_{q}))$ 
    \State $\mathcal{L} = L_{CE}(\hat{y_h}, y) + L_{Cons} + \lambda L_{CE}(\hat{y_c},y)$
    \State $\theta_H \leftarrow \theta_H - \eta \nabla_{\theta_H} \mathcal{L}$
    \State $\theta_N \leftarrow \alpha\theta_N + (1 - \alpha) \theta_H$
\EndWhile
\end{algorithmic}
\end{algorithm}

\subsection{Neocortex-Hippocampus Systems Consolidation dual-Network}

\textbf{Memory Pre-storation based Self-supervision Pretraining.}
Recent works that aimed at few-shot issues tend to leverage the prior knowledge stored in previous datasets instead of training from scratch \cite{hiller2022rethinking,hao2023class}, which is similar to the memory mechanism of human intelligence. To this end, we build a `memory pre-storation' by self-supervised pretraining.

To be specific, our approach is built around the concept of Masked Image Modeling (MIM) \cite{bao2021beit,zhou2022image}, which serves as a pretext task for self-supervised training of Vision Transformers. MIM randomly masks certain patch embeddings (tokens) with the objective of reconstructing them based on the remaining image information. The token constraints introduced assist our Transformer backbone in acquiring an embedding space that produces semantically meaningful representations for every image patch. We then fine-tune the pretrained backbone using the brain-like generalization-optimized systems consolidation process described in the following sections.

\textbf{Systems Consolidation dual-Network.}
To obtain semantically relevant information in complex background, we design a dual network that includes a Hippocampus model that executes short-term adaptation, which then gradually consolidated to the Neocortex model for structured information.
First of all, as illustrated in Figure \ref{fig:framework}, we divide the images $\boldsymbol{x} \in {\mathbb{R}^{H \times W \times C}}$ into a sequence of $M = H\times W / p^2$ patches, $p = \{p^{i}\}_{i=1}^M$, each patch $p^i\in\mathbb{R}^{{p^2}\times{C}}$. Next, we flatten and pass all the patches of the support set and query set through linear projection, the projected patches then concatenated with a [cls]\footnote{Extra learnable [class] embedding in vision transformer, representing the embedding of the whole image.} embedding as input to the ViT architecture of both the Hippocampus model ${f}\left(.;\theta_{H}\right)$ and Neocortex model ${f}\left(.;\theta_{N}\right)$. In Hippocampus model, obtaining the set of support patch embeddings $Z_s$ = ${f}\left(P_{s};\theta_{H}\right)$ with $Z_s= \{z_{s}^{nk} | n = 1, \dots, N, k = 1,\dots, K\}$, $z_{s}^{nk}= \{z_{s}^{nkm} | m = 1, \dots, M;z_{s}^{nkm}\in\mathbb{R}^D\}$and query patch embeddings $z_{q}=f_\theta\left(p_q;\theta_H\right)$ with $Z_{q}= \{z_{q}^m | m = 1, \dots, M;z_{q}\in\mathbb{R}^D\}$. 

For each of the model, we establish the similarity matrix $S$ by calculating pair-wise similarity between $z_s$ and $z_q$,  where $S\in\mathbb{R}^{N\cdot K \cdot M\times M}$, each element in $S$ is denoted as $s_{nk}^{i,j}$, acquired through Eq.(\ref{sim}):
\begin{equation}\label{sim}
    s_{nk}^{i,j}=\frac{z_s^{nki}\cdot z_q^{j}}{|z_s^{nki}|\cdot|z_q^{j}|},
\end{equation}
where $i= 1,\dots,M$ and $j=1,\dots,M$. A block-diagonal masking is employed to prevent classifying the image itself. We temperature-scale the adapted similarity logits with $1/\tau$ and aggregate the token similarity values across all elements belonging to the same support set class via a LogSumExp operation, \ie aggregating logits per class followed by a softmax – resulting in the final class prediction $\hat{y_h}$ for the query sample as Eq.(\ref{logits}). The prediction logits of Neocortex model $\hat{y_n}$ is acquired through the same calculation process.

\begin{equation}\label{logits}
 \hat{y_h} = \text{softmax} \left( \left\{ \log \sum_{k=1}^K \sum_{i=1}^{M}\sum_{j=1}^{M}exp\left(\frac{s_{nk}^{i,j}}{\tau}\right) \right\}_{n=1}^N \right)
\end{equation}

The logits from Neocortex model are then enforced a consistency loss on Hippocampus model to prevent it deviating from previous experiences, as shown in Eq.(\ref{eq:Lcons}):
\begin{equation} \label{eq:Lcons}
    L_{Cons} = \frac{\sum_{n=1}^N\left(\hat{y_h}-\hat{y_n}\right)^2}{N}
\end{equation}

To note that the Neocortex model also executes memory regulation, which leads to a `CLS' token loss $L_{CLS}$. The consistency loss between the logits of Hippocampus model and Neocortex model and cross entropy loss of Hippocampus model with a weighted sum of $L_{CLS}$ together impose a gradient to the Hippocampus model, as illustrated in Eq.(\ref{eq:LossEq}) and Eq.(\ref{eq:Hweight}), 


\begin{equation} \label{eq:LossEq}
\mathcal{L} = L_{CE}(\hat{y_h}, y) + L_{Cons} +\lambda L_{CLS}(\hat{y_c},y),
\end{equation}

\begin{equation} \label{eq:Hweight}
    \theta_H = \theta_H - \eta \nabla_{\theta_H} \mathcal{L},
\end{equation}
$\lambda$ here is a parameter to be optimized during training, which we found to work particularly well when initialized to 0.2. Finally, the Neocortex model executes the function of systems consolidation from Hippocampus model, which means the parameters are updated by taking an exponential moving average of the Hippocampus model's weights in Eq.(\ref{eq:NeocortexUpdate}), here $\alpha$ is the decay parameter.

\begin{equation} \label{eq:NeocortexUpdate}
\theta_{N}=\alpha\theta_{N}+(1-\alpha)\theta_{H}
\end{equation}

In the testing phase, we use only the Neocortex model for evaluation as it stores consolidated knowledge and structural information. 

\begin{table*}[t]
\caption{Average test accuracy on miniImageNet and tieredImagenet, for 1-shot and 5-shot scenarios, the reported results are 95\% confidence interval and tested using the established protocols.}
\label{tab:comparison1}
\begin{tabular}{@{}l l l c c c c@{}}
    \toprule
    \textbf{Model} & \textbf{Backbone} & \textbf{$\approx$ \# Params} & \multicolumn{2}{c}{\textbf{miniImageNet}} & \multicolumn{2}{c}{\textbf{tieredImageNet}} \\
                   &                   &                    & \textbf{1-shot}    & \textbf{5-shot}     & \textbf{1-shot}    & \textbf{5-shot}     \\ 
    \midrule
    SetFeat \cite{afrasiyabi2022matching}    & SetFeat-12        & 12.3 M             & 68.32$\pm$0.62         & 82.71$\pm$0.46          & 73.63$\pm$0.88          & 87.59$\pm$0.57          \\
    DeepEMD \cite{zhang2020deepemd}   & ResNet-12         & 12.4 M             & 65.91$\pm$0.82         & 82.41$\pm$0.56          & 71.16$\pm$0.87         & 86.03$\pm$0.58          \\
MELR \cite{fei2021melr}      & ResNet-12         & 12.4 M             & 67.40$\pm$0.43          & 83.40$\pm$0.28          & 72.14$\pm$0.51         & 87.01$\pm$0.35          \\
FRN \cite{wertheimer2021few}       & ResNet-12         & 12.4 M             & 66.45$\pm$0.19         & 82.83$\pm$0.13          & 72.06$\pm$0.22         & 86.89$\pm$0.14          \\

infoPatch \cite{liu2021learning}      & ResNet-12         & 12.4 M             & 67.67$\pm$0.45         & 82.44$\pm$0.31          & -         & -          \\
Mata-NVG \cite{zhang2021meta}   & ResNet-12         & 12.4 M             & 67.14$\pm$0.80         & 83.82$\pm$0.51          & 74.58$\pm$0.88         & 86.73$\pm$0.61          \\
COSOC \cite{luo2021rectifying}     & ResNet-12         & 12.4 M             & 69.28$\pm$0.49          & 85.16$\pm$0.42          & 73.57$\pm$0.43         & 87.57$\pm$0.10          \\
Meta DeepBDC \cite{xie2022joint}     & ResNet-12         & 12.4 M             & 67.34$\pm$0.43         & 84.46$\pm$0.28          & 72.34$\pm$0.49         & 87.31$\pm$0.32         \\
AMMMD \cite{wu2024ammd} & ResNet-12  & 12.4 M &  70.31 ± 0.45& 85.22 ± 0.29 &	74.22 ± 0.50	& 87.55 ± 0.34\\
\midrule

FEAT \cite{ye2020few}      & WRN-28-10         & 36.5 M             & 65.10$\pm$0.20         & 81.11$\pm$0.14          & 70.41$\pm$0.23         & 84.38$\pm$0.16          \\
MetaQDA \cite{zhang2021shallow}      & WRN-28-10         & 36.5 M             & 67.83$\pm$0.64         & 84.28$\pm$0.69          & 74.33$\pm$0.65         & 89.56$\pm$0.79          \\
\midrule
    FewTURE \cite{hiller2022rethinking}   & ViT-S/16          & 22 M               & 68.02$\pm$0.88         & 84.51$\pm$0.53          & 72.96$\pm$0.92         & 86.43$\pm$0.67          \\
    CPEA \cite{hao2023class} & ViT-S/16 & 22 M & 71.97$\pm$0.65 & 87.06$\pm$0.38 & 76.93$\pm$0.70 & \underline{90.12$\pm$0.45}\\
    AMMD \cite{wu2024ammd} & ViT-S/16 & 22 M & \underline{73.70$\pm$0.44}	&   \underline{88.12$\pm$0.25}	& \underline{77.59$\pm$0.50}	& \textbf{90.38$\pm$0.30}\\
    \textbf{SCAM-Net (ours)} & ViT-S/16  & 22 M & \textbf{75.93$\pm$0.45} & \textbf{89.75$\pm$0.22} & \textbf{78.89$\pm$0.52} & 88.67$\pm$0.65 \\
    \bottomrule
\end{tabular}
\end{table*}

\subsection{Generalization-optimized Adaptive Memory Regulation.}
To deal with intra-class data drifting conditions, the Neocortex maintains a long-term memory for adaptive memory regulation, as illustrated in Figure \ref{fig:framework}. The CLS token, according to \cite{dosovitskiy2020image} serve as image's global representation, which we store in the long-term memory with the image's class label as key. In each task, the CLS vector $C = \{c^{nk}|n = 1,\dots,N,k = 1,\dots,K\}$ firstly average on the second dimension to get $C' = \{c^n|n = 1,\dots,N\}$, then look up in the long-term memory to see if corresponding class exist, if not, append the current memory with the class as key and $C'$ as value; if it exists, then average between the existing value and $C'$, as shown in Eq.(\ref{eq:cls1}). The ${C}_q$ denotes the CLS tokens of query set, ${L}$ is the corresponding class label. After we obtain regulated CLS tokens of support set $C_{opt}$, we calculate similarity between them and the query CLS tokens which is denoted as $\hat{y_c}$, as a prediction of image's global representation, as illustrated in Eq.(\ref{eq:cls3}). 

\begin{equation}\label{eq:cls1}
    \begin{aligned}
        \text{if } {L} &\notin {M}: \quad {M}[{L}] = C' \\
        \text{else:} \quad &{M}[{L}] = \frac{1}{2} \left( {M}[{L}] + C' \right)
    \end{aligned}
\end{equation}

\begin{equation}\label{eq:cls3}
    \hat{y_c} = \text{softmax}\left( \frac{{C_{opt}}\cdot {C}_q}{|C_{opt}|\cdot|{C}_q|}\right)
\end{equation}

\section{Experiments}
\begin{table*} 
\caption{Average test accuracy on FC100,  for 1-shot and 5-shot scenarios, the reported results are 95\% confidence interval and tested using the established protocols.}
\label{tab:comparison2}
\begin{tabular}{@{}llc c c c c@{}}  
\toprule
\textbf{Model}   & \textbf{Backbone} & \textbf{$\approx$ \# Params} & \multicolumn{2}{c}{\textbf{CIFAR-FS}} & \multicolumn{2}{c}{\textbf{FC100}}                  \\ 
                 &                   &                              & \textbf{1-shot}    & \textbf{5-shot} & \textbf{1-shot}          & \textbf{5-shot}         \\ 
\midrule

MetaOpt \cite{lee2019meta}     & ResNet-12         & 12.4 M    
& 72.60$\pm$0.70 & 84.30$\pm$0.50 & 41.10$\pm$0.60           & 55.30$\pm$0.60           \\
MABAS \cite{kim2020model}     & ResNet-12         & 12.4 M                       & 73.51±0.92&85.65$\pm$0.65& 42.31$\pm$0.75           & 58.16$\pm$0.78           \\
RFS \cite{tian2020rethinking}         & ResNet-12         & 12.4 M                &73.90$\pm$0.80 & 86.90$\pm$0.50       & 44.60$\pm$0.70           & 60.90$\pm$0.60           \\
Meta-NVG \cite{zhang2021meta}    & ResNet-12         & 12.4 M                     & 74.63$\pm$0.91&86.45$\pm$0.59 & 46.40$\pm$0.81           & 61.33$\pm$0.71           \\
TPMN \cite{wu2021task}        & ResNet-12         & 12.4 M                       & 75.50$\pm$0.90& 87.20$\pm$0.60 & 46.93$\pm$0.71           & 63.26$\pm$0.74           \\

\midrule
FewTURE \cite{hiller2022rethinking}     & ViT-S/16          & 22 M                  &76.10$\pm$0.88 & 86.14$\pm$0.64       & 46.20$\pm$0.79           & 63.14$\pm$0.73           \\
CPEA \cite{hao2023class}            & ViT-S/16          & 22 M                      & 77.82$\pm$0.66 & 88.98$\pm$0.45   & 47.24$\pm$0.58           & 65.02$\pm$0.60           \\
 AMMD \cite{wu2024ammd} & ViT-S/16 & 22 M & \underline{78.10$\pm$0.47} & \underline{89.66$\pm$0.31} & \underline{47.30$\pm$0.41}& \textbf{66.41$\pm$0.41}\\
\textbf{SCAM-Net (ours)}&ViT-S/16  & 22 M & \textbf{79.45$\pm$0.53}& \textbf{90.90$\pm$0.45}&\textbf{47.48$\pm$0.32} & \underline{65.85$\pm$0.75} \\
\bottomrule
\end{tabular}
\end{table*}

\subsection{Experimental Settings}

\textbf{Datasets.} Following existing works, we evaluate our method on four few-shot classification benchmark datasets, \ie, miniImagenet \cite{vinyals2016matching}, tieredImagenet \cite{ren2018meta}, CIFAR-FS and FC-100 \cite{oreshkin2018tadam}. We followed the standards of previous research \cite{ye2020few} to split each dataset into training/validation/test datasets. Note that the training set, validation set and test set are split without overlap.

\textbf{Backbone.} We use the monolithic ViT architecture in its `small' form \cite{dosovitskiy2020image} as backbone, since \cite{hiller2022rethinking} demonstrates that Transformer-only architectures in conjunction with self-supervised pretraining can be successfully used in few-shot settings without the need of convolutional backbones or any additional data. The model receives images with resolution of $224 \times 224$ and the patch embedding dimension is 384-d.

\textbf{Implementation Details.}
The training procedure consists of two stages. In the first stage, We employ the training strategy of \cite{zhou2022image} to pretrain our model and sticked to the hyperparameter settings reported in their work. The model was trained with a batch size of 512 for 800 and 1600 epoches with four Nvidia A100 GPUs with 40GB capacity. Note that only the training set of each datasets was used for pretraining. In the second stage, we mimic the Hippocampus-Neocortex dual net to conduct training on episodes that sampled from the training classes. We use SGD as the optimizer. The initial learning rate of Hippocampus model is set to be 0.0002 by default, the Neocortex model is updated with decay parameter $\alpha$ that initially set as 0.99. $\lambda$ is the weight parameter of CLS token similarity, which we found that initialized as 0.2 works well on most of the datasets, which is also optimized during training. 

\textbf{Evaluation Protocol.}
We conduct experiments on 5-way 1-shot and 5-way 5-shot classification tasks, and evaluate at each epoch on 600 randomly sampled episodes from the validation set. During test time, we use Neocortex model to test the final performance with 15 query images per class on 2500 test episodes.

\subsection{Comparison Results}
Table \ref{tab:comparison1} and Table \ref{tab:comparison2} show that SCAM-Net outperforms existing counterparts that lack biological plausibility, demonstrating the effectiveness of the generalization-optimized systems consolidation mechanism.  From technological perspective, the slow consolidation process of Neocortex model retains structural knowledge which efficiently alleviate the problem of data drift, and inducing the adaptive memory regulation the support images made the samples more closely attributed around clustering center, improving the reliability of support images.

\subsection{Ablation Study}
The key components that build the process of generalization-optimized systems consolidation are ablated, and the experiments are conducted on miniImagenet \cite{vinyals2016matching}.

\textbf{Systems Consolidation.} 
We implement systems consolidation with the goal of enabling the model to extract meaningful features in data drifting conditions. The short-term adaptation captures task-specific nuances and long-term memory consolidation preserves global knowledge across tasks. We can tell from Table \ref{tab:system} that systems consolidation, \eg the interplay of Hippocampus-Neocortex dual-network enhanced the performance at a large scale for it ultimately reduces background interference and focuses on the semantically relevant regions.

\begin{table}
\caption{Impact of systems consolidation on few-shot image classification performance.}
\label{tab:system}
\begin{tabular}{ccl}
\toprule    
Systems consolidation & 1-shot  &  5-shot \\ 
\midrule 
$\times$ &  66.8$\pm$0.45 & 83.40$\pm$0.33  \\ 
\checkmark &  72.19$\pm$0.30 & 85.24$\pm$0.25 \\
\bottomrule 
\end{tabular}
\end{table}

\begin{figure}[htb]
    \centering
    \includegraphics[width=0.9\linewidth]{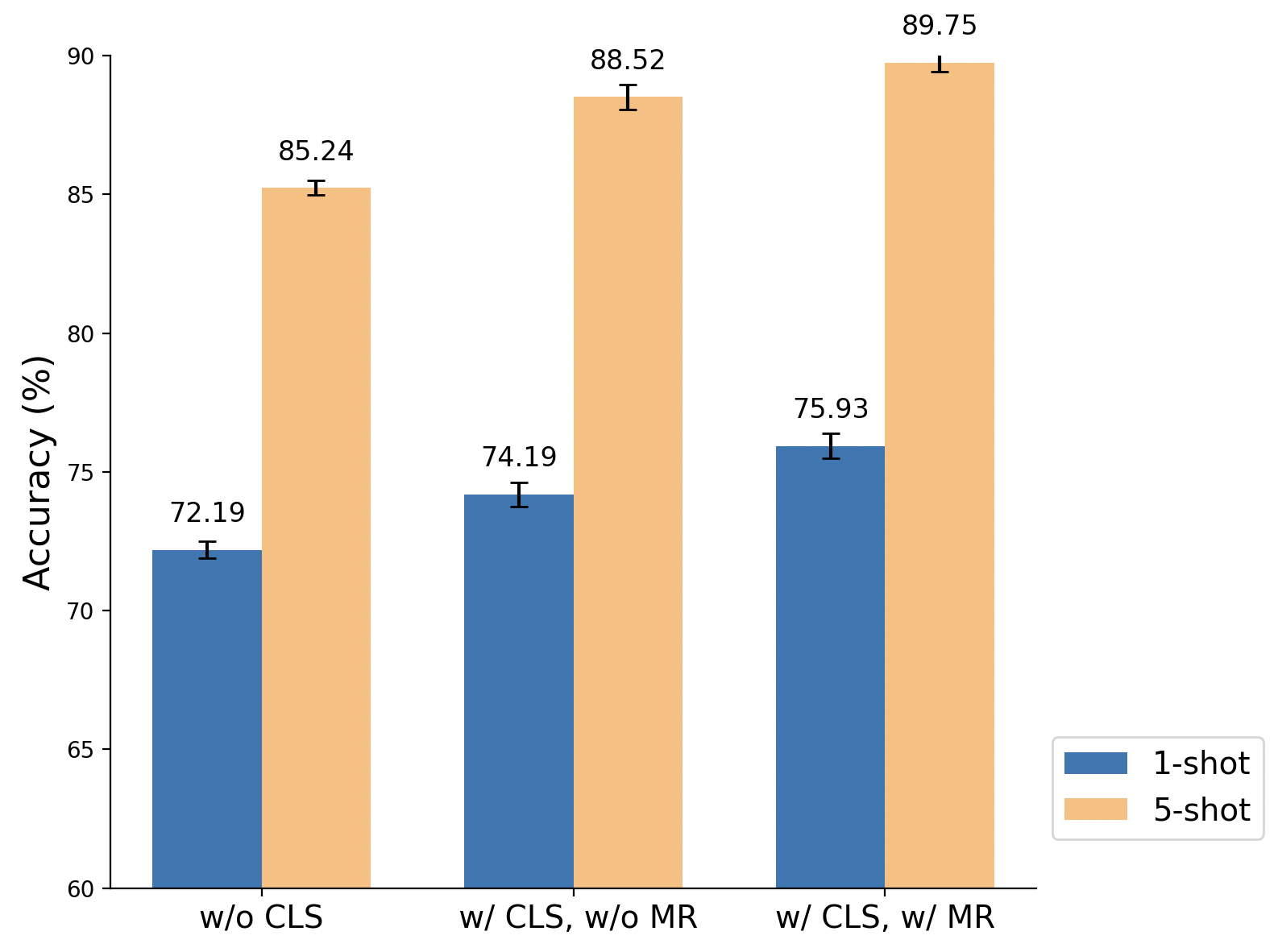}
    \caption{Impact of adaptive memory regulation (MR) on few-shot image classification performance.}
    \label{MR}
\end{figure}

\textbf{Generalization-optimized Adaptive Memory Regulation (MR).}
We model the process of optimizing generalization by simulating a long-term memory inside Neocortex model, in which maintains the CLS token that represents the corresponding class, and CLS tokens of support sets can be regulated base on the class information stored in long-term memory. We conduct experiments on with/without CLS token, and under `with' condition, with/without memory regulation. As shown in Figure \ref{MR}, on 1-shot setting. After the image representation (CLS token) was considered, the overall performance was enhanced by 2\%, and after we adopted memory regulation, it is further promoted by 1.74\%, demonstrating that memory regulation significantly enhances adaptability and reliability of support set tokens.

\begin{figure*}[t]
  \centering
  \includegraphics[width=0.9\linewidth]{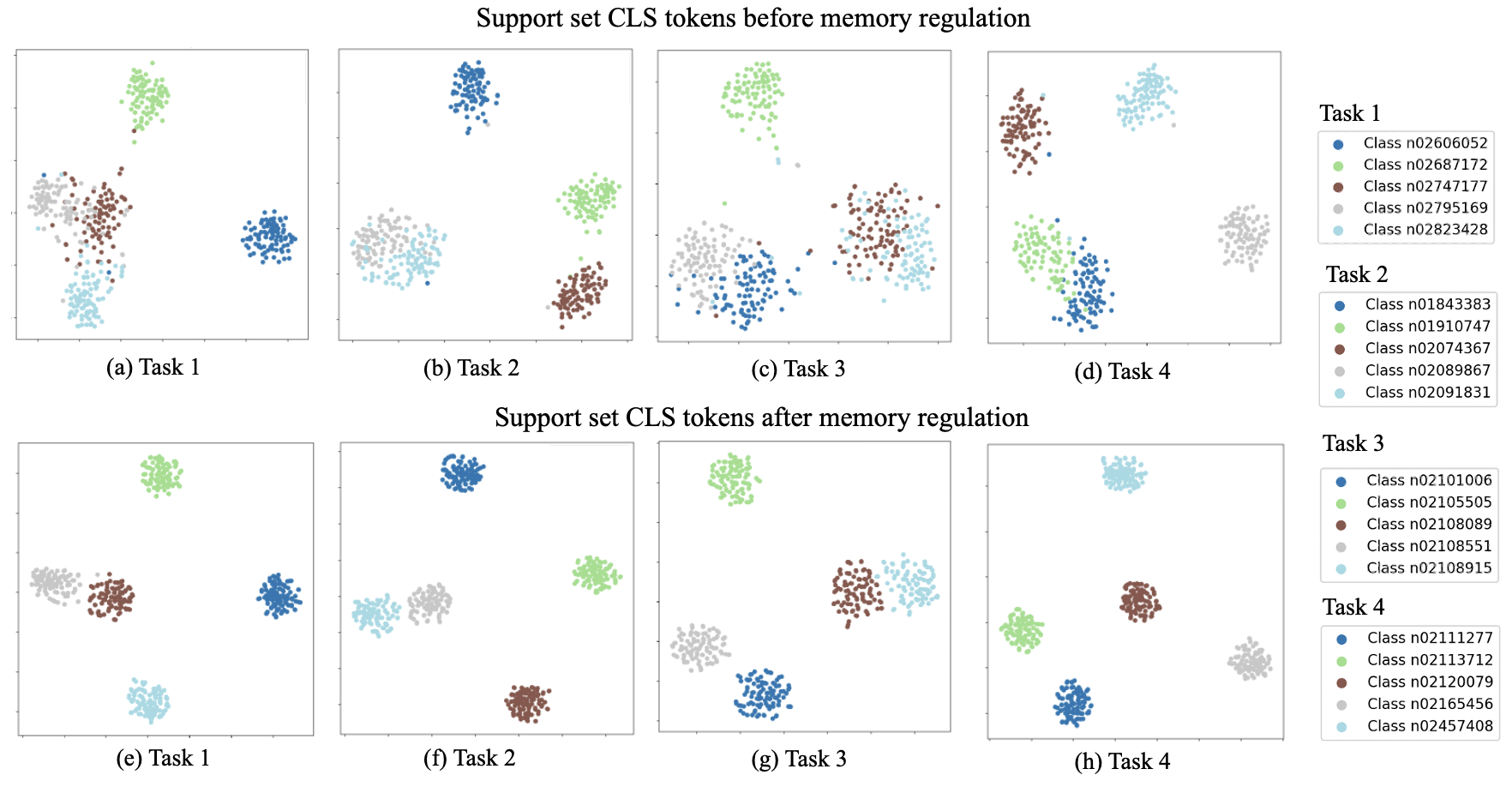}
  \caption{Visualization of support set CLS tokens before/after adaptive memory regulation of four randomly sampled 5-way 1-shot classification tasks. (a), (b), (c) and (d) show the visualization results before memory regulation. (e), (f), (g) and (h) show the corresponding results after memory regulation. Adaptive memory make the image's representation of certain class aggregated, hence the information of support set is less biased.}
  \label{fig:visualization}
\end{figure*}

\textbf{Visualization of Memory Regulation.}
Figure \ref{fig:visualization} (a), (b), (c) and (d) depicts support set CLS tokens before memory regulation, and (e), (f), (g) and (h) shows the corresponding results after memory regulation. To exclude the effect of parameter optimization, we didn't perform any gradient descent during the process. As can be seen from Figure \ref{fig:visualization}, the generalization-optimized adaptive memory regulation made the scattered dots aggregated to their center, due to the complexity of real-world images, the CLS token that represents the entire image can disperse across feature spaces of all classes, even over lap with other classes, as shown in Figure \ref{fig:visualization} (c) Task 3. After the regulation with history information, particular deviant CLS token is being pulled back, making the CLS tokens of support set less biased and result in a clustered form. 

\begin{table}
    \caption{Comparison of training speed using single and dual network on miniImagenet on the same A800 GPU.}
    \centering
    \begin{tabular}{ccc}
    \toprule
       &   single net  &   dual net  \\
    \midrule 
    training speed [it/s] & 4.18 & 2.68\\
    \bottomrule
    \end{tabular}
    \label{tab:my_label}
\end{table}
\textbf{Training Speed.}
The dual network structure will increase the training time since more trainable parameters are included. Therefore, we listed a comparison of training speed between single network and dual network on the same A800 GPU. As shown in Table \ref{tab:my_label}, our dual network retains structured information at slightly increased computational cost. During inference, we used only one Neocortex network so there won't be additional computational cost.
\subsection{Hyperparameter Sensitivity}
\textbf{Consistency Loss.}
We conducted experiments on the loss function used for restriction between Hippocampus model and Neocortex model. Table \ref{tab:cont} shows that MSE loss performs slightly better than KL loss. We conclude that since MSE directly penalizes the difference in predictions, promoting more stable and exact matching between the Hippocampus and Neocortex models, which is crucial in scenarios with limited data. Meanwhile KL divergence, while effective at aligning distributions, might be less straightforward in ensuring precise agreement when limited samples provided.
\begin{table}
    \centering
    \caption{Choices of consistency loss between dual-network.}
    \label{tab:cont}
    \begin{tabular}{ccc}
    \toprule      
    Consistency loss & 1-shot  &  5-shot \\ 
    \midrule 
    MSE loss  & 75.93 $\pm$ 0.45 &  89.75 $\pm$ 0.22 \\  
    KL loss & 75.60 $\pm$ 0.81 & 88.33 $\pm$ 0.52\\
    \bottomrule 
    \end{tabular} 
\end{table}

\textbf{Distance Metrics.}
We also conduct experiments on distance metrics used in Eq.(\ref{eq:cls3}), in which was Cosine distance by default. Table \ref{tab:distance} shows that both Cosine and Manhattan distance can boost performance, Euclid distance seems less effective. Cosine and Manhattan distances are less sensitive to variations in vector magnitude and are more robust to outliers. Cosine distance measures angular differences, focusing on direction rather than magnitude, which can be particularly effective in high-dimensional spaces where Euclidean distance may not capture the true dissimilarities due to scale sensitivity.

\begin{table}
\centering
\caption{Impact of choices in Eq.(\ref{eq:cls3}) on the few-shot classification performance.}
\label{tab:distance}
\begin{tabular}{ccc}
\toprule 
Choices in Eq.(\ref{eq:cls3}) & 1-shot  &  5-shot \\ 
\midrule  
Cosine &  75.93 $\pm$ 0.45 &  89.75 $\pm$ 0.22  \\ 
Euclid &  74.92 $\pm$ 0.53 & 88.60 $\pm$ 0.47  \\
Manhattan &  75.22 $\pm$ 0.54 & 89.24 $\pm$ 0.45 \\
\bottomrule     
\end{tabular}
\end{table}

\section{Conclusion}
In this paper, we propose a generalization-optimized Systems Consolidation Adaptive Memory dual-Network (SCAM-Net) to address supervision collapse problem in few-shot image classification. SCAM-Net imitates the biological structure of human's complementary learning system by constructing a dual network structure that consolidates structured information. To alleviate the negative effect of data drift, we considered the optimize generalization principle in systems consolidation by storing images' global representation in long-term memory which adaptively regulated based on prior knowledge. This approach successfully addresses the difficulty of identifying semantically relevant feature and minimizes the bias of support set features therefore improves the reliability of sample images. Ablation studies have proved the validity of each component and visualization of memory regulation demonstrate its absolute necessity. Comparison experiments on benchmark datasets proved that our SCAM-Net outperforms counterparts while achieving state-of-the-art performance.

\bibliographystyle{named}
\bibliography{ijcai24}

\begin{thebibliography}{}

\bibitem[\protect\citeauthoryear{Afrasiyabi \bgroup \em et al.\egroup }{2022}]{afrasiyabi2022matching}
Arman Afrasiyabi, Hugo Larochelle, Jean-Fran{\c{c}}ois Lalonde, and Christian Gagn{\'e}.
\newblock Matching feature sets for few-shot image classification.
\newblock In {\em Proceedings of the IEEE/CVF conference on computer vision and pattern recognition}, pages 9014--9024, 2022.

\bibitem[\protect\citeauthoryear{Arani \bgroup \em et al.\egroup }{2022}]{aranilearning}
Elahe Arani, Fahad Sarfraz, and Bahram Zonooz.
\newblock Learning fast, learning slow: A general continual learning method based on complementary learning system.
\newblock In {\em International Conference on Learning Representations}, 2022.

\bibitem[\protect\citeauthoryear{Atallah \bgroup \em et al.\egroup }{2004}]{atallah2004Hippocampus}
Hisham~E Atallah, Michael~J Frank, and Randall~C O'Reilly.
\newblock Hippocampus, cortex, and basal ganglia: Insights from computational models of complementary learning systems.
\newblock {\em Neurobiology of learning and memory}, 82(3):253--267, 2004.

\bibitem[\protect\citeauthoryear{Bao \bgroup \em et al.\egroup }{2021}]{bao2021beit}
Hangbo Bao, Li~Dong, Songhao Piao, and Furu Wei.
\newblock Beit: Bert pre-training of image transformers.
\newblock {\em arXiv preprint arXiv:2106.08254}, 2021.

\bibitem[\protect\citeauthoryear{Doersch \bgroup \em et al.\egroup }{2020}]{doersch2020crosstransformers}
Carl Doersch, Ankush Gupta, and Andrew Zisserman.
\newblock Crosstransformers: spatially-aware few-shot transfer.
\newblock {\em Advances in Neural Information Processing Systems}, 33:21981--21993, 2020.

\bibitem[\protect\citeauthoryear{Dosovitskiy}{2020}]{dosovitskiy2020image}
Alexey Dosovitskiy.
\newblock An image is worth 16x16 words: Transformers for image recognition at scale.
\newblock {\em arXiv preprint arXiv:2010.11929}, 2020.

\bibitem[\protect\citeauthoryear{Duan \bgroup \em et al.\egroup }{2024}]{duan2024few}
Zhibin Duan, Zhiyi Lv, Chaojie Wang, Bo~Chen, Bo~An, and Mingyuan Zhou.
\newblock Few-shot generation via recalling brain-inspired episodic-semantic memory.
\newblock {\em Advances in Neural Information Processing Systems}, 36, 2024.

\bibitem[\protect\citeauthoryear{Fei \bgroup \em et al.\egroup }{2021}]{fei2021melr}
Nanyi Fei, Zhiwu Lu, Tao Xiang, and Songfang Huang.
\newblock Melr: Meta-learning via modeling episode-level relationships for few-shot learning.
\newblock In {\em International Conference on Learning Representations}, 2021.

\bibitem[\protect\citeauthoryear{Finn \bgroup \em et al.\egroup }{2017}]{finn2017model}
Chelsea Finn, Pieter Abbeel, and Sergey Levine.
\newblock Model-agnostic meta-learning for fast adaptation of deep networks.
\newblock In {\em International conference on machine learning}, pages 1126--1135. PMLR, 2017.

\bibitem[\protect\citeauthoryear{French}{1999}]{french1999catastrophic}
Robert~M French.
\newblock Catastrophic forgetting in connectionist networks.
\newblock {\em Trends in cognitive sciences}, 3(4):128--135, 1999.

\bibitem[\protect\citeauthoryear{Han \bgroup \em et al.\egroup }{2024}]{han2024novel}
Yuyang Han, Xiuxing Li, Tianyuan Jia, Qixin Wang, Chaoqiong Fan, and Xia Wu.
\newblock A novel sleep mechanism inspired continual learning algorithm.
\newblock {\em Guidance, Navigation and Control}, page 2441003, 2024.

\bibitem[\protect\citeauthoryear{Hao \bgroup \em et al.\egroup }{2023}]{hao2023class}
Fusheng Hao, Fengxiang He, Liu Liu, Fuxiang Wu, Dacheng Tao, and Jun Cheng.
\newblock Class-aware patch embedding adaptation for few-shot image classification.
\newblock In {\em Proceedings of the IEEE/CVF International Conference on Computer Vision}, pages 18905--18915, 2023.

\bibitem[\protect\citeauthoryear{Hiller \bgroup \em et al.\egroup }{2022}]{hiller2022rethinking}
Markus Hiller, Rongkai Ma, Mehrtash Harandi, and Tom Drummond.
\newblock Rethinking generalization in few-shot classification.
\newblock {\em Advances in Neural Information Processing Systems}, 35:3582--3595, 2022.

\bibitem[\protect\citeauthoryear{Hou \bgroup \em et al.\egroup }{2019}]{hou2019cross}
Ruibing Hou, Hong Chang, Bingpeng Ma, Shiguang Shan, and Xilin Chen.
\newblock Cross attention network for few-shot classification.
\newblock {\em Advances in neural information processing systems}, 32, 2019.

\bibitem[\protect\citeauthoryear{Kim \bgroup \em et al.\egroup }{2020}]{kim2020model}
Jaekyeom Kim, Hyoungseok Kim, and Gunhee Kim.
\newblock Model-agnostic boundary-adversarial sampling for test-time generalization in few-shot learning.
\newblock In {\em Computer Vision--ECCV 2020: 16th European Conference, Glasgow, UK, August 23--28, 2020, Proceedings, Part I 16}, pages 599--617. Springer, 2020.

\bibitem[\protect\citeauthoryear{Lee \bgroup \em et al.\egroup }{2019}]{lee2019meta}
Kwonjoon Lee, Subhransu Maji, Avinash Ravichandran, and Stefano Soatto.
\newblock Meta-learning with differentiable convex optimization.
\newblock In {\em Proceedings of the IEEE/CVF conference on computer vision and pattern recognition}, pages 10657--10665, 2019.

\bibitem[\protect\citeauthoryear{Leng \bgroup \em et al.\egroup }{2024}]{leng2024meta}
Zhixiong Leng, Maofa Wang, Quan Wan, Yanlin Xu, Bingchen Yan, and Shaohua Sun.
\newblock Meta-learning of feature distribution alignment for enhanced feature sharing.
\newblock {\em Knowledge-Based Systems}, 296:111875, 2024.

\bibitem[\protect\citeauthoryear{Liu \bgroup \em et al.\egroup }{2021}]{liu2021learning}
Chen Liu, Yanwei Fu, Chengming Xu, Siqian Yang, Jilin Li, Chengjie Wang, and Li~Zhang.
\newblock Learning a few-shot embedding model with contrastive learning.
\newblock In {\em Proceedings of the AAAI conference on artificial intelligence}, volume~35, pages 8635--8643, 2021.

\bibitem[\protect\citeauthoryear{Liu \bgroup \em et al.\egroup }{2022}]{liu2022learning}
Yang Liu, Weifeng Zhang, Chao Xiang, Tu~Zheng, Deng Cai, and Xiaofei He.
\newblock Learning to affiliate: Mutual centralized learning for few-shot classification.
\newblock In {\em Proceedings of the IEEE/CVF conference on computer vision and pattern recognition}, pages 14411--14420, 2022.

\bibitem[\protect\citeauthoryear{Luo \bgroup \em et al.\egroup }{2021}]{luo2021rectifying}
Xu~Luo, Longhui Wei, Liangjian Wen, Jinrong Yang, Lingxi Xie, Zenglin Xu, and Qi~Tian.
\newblock Rectifying the shortcut learning of background for few-shot learning.
\newblock {\em Advances in Neural Information Processing Systems}, 34:13073--13085, 2021.

\bibitem[\protect\citeauthoryear{Luo \bgroup \em et al.\egroup }{2023}]{luo2023closer}
Xu~Luo, Hao Wu, Ji~Zhang, Lianli Gao, Jing Xu, and Jingkuan Song.
\newblock A closer look at few-shot classification again.
\newblock In {\em International Conference on Machine Learning}, pages 23103--23123. PMLR, 2023.

\bibitem[\protect\citeauthoryear{McClelland \bgroup \em et al.\egroup }{1995}]{mcclelland1995there}
James~L McClelland, Bruce~L McNaughton, and Randall~C O'Reilly.
\newblock Why there are complementary learning systems in the hippocampus and neocortex: insights from the successes and failures of connectionist models of learning and memory.
\newblock {\em Psychological review}, 102(3):419, 1995.

\bibitem[\protect\citeauthoryear{Nichol}{2018}]{nichol2018first}
A~Nichol.
\newblock On first-order meta-learning algorithms.
\newblock {\em arXiv preprint arXiv:1803.02999}, 2018.

\bibitem[\protect\citeauthoryear{Oreshkin \bgroup \em et al.\egroup }{2018}]{oreshkin2018tadam}
Boris Oreshkin, Pau Rodr{\'\i}guez~L{\'o}pez, and Alexandre Lacoste.
\newblock Tadam: Task dependent adaptive metric for improved few-shot learning.
\newblock {\em Advances in neural information processing systems}, 31, 2018.

\bibitem[\protect\citeauthoryear{Ren \bgroup \em et al.\egroup }{2018}]{ren2018meta}
Mengye Ren, Eleni Triantafillou, Sachin Ravi, Jake Snell, Kevin Swersky, Joshua~B Tenenbaum, Hugo Larochelle, and Richard~S Zemel.
\newblock Meta-learning for semi-supervised few-shot classification.
\newblock {\em arXiv preprint arXiv:1803.00676}, 2018.

\bibitem[\protect\citeauthoryear{Rostami \bgroup \em et al.\egroup }{2019}]{rostami2019complementary}
Mohammad Rostami, Soheil Kolouri, and Praveen~K Pilly.
\newblock Complementary learning for overcoming catastrophic forgetting using experience replay.
\newblock In {\em Proceedings of the 28th International Joint Conference on Artificial Intelligence}, pages 3339--3345, 2019.

\bibitem[\protect\citeauthoryear{Russakovsky \bgroup \em et al.\egroup }{2015}]{russakovsky2015imagenet}
Olga Russakovsky, Jia Deng, Hao Su, Jonathan Krause, Sanjeev Satheesh, Sean Ma, Zhiheng Huang, Andrej Karpathy, Aditya Khosla, Michael Bernstein, et~al.
\newblock Imagenet large scale visual recognition challenge.
\newblock {\em International journal of computer vision}, 115:211--252, 2015.

\bibitem[\protect\citeauthoryear{Schapiro \bgroup \em et al.\egroup }{2017}]{schapiro2017complementary}
Anna~C Schapiro, Nicholas~B Turk-Browne, Matthew~M Botvinick, and Kenneth~A Norman.
\newblock Complementary learning systems within the hippocampus: a neural network modelling approach to reconciling episodic memory with statistical learning.
\newblock {\em Philosophical Transactions of the Royal Society B: Biological Sciences}, 372(1711):20160049, 2017.

\bibitem[\protect\citeauthoryear{Snell \bgroup \em et al.\egroup }{2017}]{snell2017prototypical}
Jake Snell, Kevin Swersky, and Richard Zemel.
\newblock Prototypical networks for few-shot learning.
\newblock {\em Advances in neural information processing systems}, 30, 2017.

\bibitem[\protect\citeauthoryear{Sorscher \bgroup \em et al.\egroup }{2022}]{sorscher2022neural}
Ben Sorscher, Surya Ganguli, and Haim Sompolinsky.
\newblock Neural representational geometry underlies few-shot concept learning.
\newblock {\em Proceedings of the National Academy of Sciences}, 119(43):e2200800119, 2022.

\bibitem[\protect\citeauthoryear{Spens and Burgess}{2024}]{spens2024generative}
Eleanor Spens and Neil Burgess.
\newblock A generative model of memory construction and consolidation.
\newblock {\em Nature Human Behaviour}, 8(3):526--543, 2024.

\bibitem[\protect\citeauthoryear{Sun \bgroup \em et al.\egroup }{2023}]{sun2023organizing}
Weinan Sun, Madhu Advani, Nelson Spruston, Andrew Saxe, and James~E Fitzgerald.
\newblock Organizing memories for generalization in complementary learning systems.
\newblock {\em Nature neuroscience}, 26(8):1438--1448, 2023.

\bibitem[\protect\citeauthoryear{Sung \bgroup \em et al.\egroup }{2018}]{sung2018learning}
Flood Sung, Yongxin Yang, Li~Zhang, Tao Xiang, Philip~HS Torr, and Timothy~M Hospedales.
\newblock Learning to compare: Relation network for few-shot learning.
\newblock In {\em Proceedings of the IEEE conference on computer vision and pattern recognition}, pages 1199--1208, 2018.

\bibitem[\protect\citeauthoryear{Tian \bgroup \em et al.\egroup }{2020}]{tian2020rethinking}
Yonglong Tian, Yue Wang, Dilip Krishnan, Joshua~B Tenenbaum, and Phillip Isola.
\newblock Rethinking few-shot image classification: a good embedding is all you need?
\newblock In {\em Computer Vision--ECCV 2020: 16th European Conference, Glasgow, UK, August 23--28, 2020, Proceedings, Part XIV 16}, pages 266--282. Springer, 2020.

\bibitem[\protect\citeauthoryear{Vinyals \bgroup \em et al.\egroup }{2016}]{vinyals2016matching}
Oriol Vinyals, Charles Blundell, Timothy Lillicrap, Daan Wierstra, et~al.
\newblock Matching networks for one shot learning.
\newblock {\em Advances in neural information processing systems}, 29, 2016.

\bibitem[\protect\citeauthoryear{Wertheimer \bgroup \em et al.\egroup }{2021}]{wertheimer2021few}
Davis Wertheimer, Luming Tang, and Bharath Hariharan.
\newblock Few-shot classification with feature map reconstruction networks.
\newblock In {\em Proceedings of the IEEE/CVF conference on computer vision and pattern recognition}, pages 8012--8021, 2021.

\bibitem[\protect\citeauthoryear{Wu \bgroup \em et al.\egroup }{2021}]{wu2021task}
Jiamin Wu, Tianzhu Zhang, Yongdong Zhang, and Feng Wu.
\newblock Task-aware part mining network for few-shot learning.
\newblock In {\em Proceedings of the IEEE/CVF International Conference on Computer Vision}, pages 8433--8442, 2021.

\bibitem[\protect\citeauthoryear{Wu \bgroup \em et al.\egroup }{2024}]{wu2024ammd}
Ji~Wu, Shipeng Wang, and Jian Sun.
\newblock Ammd: Attentive maximum mean discrepancy for few-shot image classification.
\newblock {\em Pattern Recognition}, page 110680, 2024.

\bibitem[\protect\citeauthoryear{Xie \bgroup \em et al.\egroup }{2022}]{xie2022joint}
Jiangtao Xie, Fei Long, Jiaming Lv, Qilong Wang, and Peihua Li.
\newblock Joint distribution matters: Deep brownian distance covariance for few-shot classification.
\newblock In {\em Proceedings of the IEEE/CVF conference on computer vision and pattern recognition}, pages 7972--7981, 2022.

\bibitem[\protect\citeauthoryear{Xu \bgroup \em et al.\egroup }{2021}]{xu2021learning}
Chengming Xu, Yanwei Fu, Chen Liu, Chengjie Wang, Jilin Li, Feiyue Huang, Li~Zhang, and Xiangyang Xue.
\newblock Learning dynamic alignment via meta-filter for few-shot learning.
\newblock In {\em Proceedings of the IEEE/CVF conference on computer vision and pattern recognition}, pages 5182--5191, 2021.

\bibitem[\protect\citeauthoryear{Ye \bgroup \em et al.\egroup }{2020}]{ye2020few}
Han-Jia Ye, Hexiang Hu, De-Chuan Zhan, and Fei Sha.
\newblock Few-shot learning via embedding adaptation with set-to-set functions.
\newblock In {\em Proceedings of the IEEE/CVF conference on computer vision and pattern recognition}, pages 8808--8817, 2020.

\bibitem[\protect\citeauthoryear{Zhang \bgroup \em et al.\egroup }{2020}]{zhang2020deepemd}
Chi Zhang, Yujun Cai, Guosheng Lin, and Chunhua Shen.
\newblock Deepemd: Few-shot image classification with differentiable earth mover's distance and structured classifiers.
\newblock In {\em Proceedings of the IEEE/CVF conference on computer vision and pattern recognition}, pages 12203--12213, 2020.

\bibitem[\protect\citeauthoryear{Zhang \bgroup \em et al.\egroup }{2021a}]{zhang2021meta}
Chi Zhang, Henghui Ding, Guosheng Lin, Ruibo Li, Changhu Wang, and Chunhua Shen.
\newblock Meta navigator: Search for a good adaptation policy for few-shot learning.
\newblock In {\em Proceedings of the IEEE/CVF international conference on computer vision}, pages 9435--9444, 2021.

\bibitem[\protect\citeauthoryear{Zhang \bgroup \em et al.\egroup }{2021b}]{zhang2021shallow}
Xueting Zhang, Debin Meng, Henry Gouk, and Timothy~M Hospedales.
\newblock Shallow bayesian meta learning for real-world few-shot recognition.
\newblock In {\em Proceedings of the IEEE/CVF international conference on computer vision}, pages 651--660, 2021.

\bibitem[\protect\citeauthoryear{Zhou \bgroup \em et al.\egroup }{2022}]{zhou2022image}
Jinghao Zhou, Chen Wei, Huiyu Wang, Wei Shen, Cihang Xie, Alan Yuille, and Tao Kong.
\newblock Image bert pre-training with online tokenizer.
\newblock In {\em International Conference on Learning Representations}, 2022.

\end{thebibliography}

\end{document}